%% file: main.tex
\documentclass[11pt]{article} 
\usepackage[margin=1in]{geometry}
\usepackage{times}
\usepackage{microtype}
\usepackage{wrapfig}
\usepackage{booktabs}
\usepackage{graphicx}
\usepackage{amssymb}
\usepackage{amsmath}
\usepackage{comment}
\usepackage{xcolor}
\usepackage{natbib}
\usepackage{authblk}

\input{math_commands.tex}

\usepackage{hyperref}
\usepackage{url}

\title{ResDiffFRG: Residual Diffusion for Multiple Appropriate Facial Reaction Generation}

\author[1]{Shizhe Liu}
\author[2]{Jiayan Gu}
\author[3]{Xiangyu Kong}
\author[3]{Siyang Song}

\affil[1]{Department of Computer Science, University of Oxford, Oxford, United Kingdom}
\affil[2]{School of Artificial Intelligence and Big Data, Hefei University, Hefei, China}
\affil[3]{Department of Computer Science, University of Exeter, Exeter, United Kingdom}

\date{}

\begin{document}

\maketitle

\begin{abstract}

In dyadic human speaker-listener conversations, the listener's facial reactions play an important role in ensuring that the speaker accurately perceives the listener's emotional states. Since human facial reactions are inherently non-deterministic, the ability to generate multiple appropriate human-like facial reactions is crucial for realistic and engaging human-agent interactions. 
Although diffusion models are naturally suited to such one-to-many generation, existing diffusion-based Multiple Appropriate Facial Reaction Generation (MAFRG) methods attempt to denoise random Gaussian initialisations directly into multiple appropriate facial reactions (AFRs), conditioned on the perceived speaker behaviour. 
These random initialisations are usually not well-aligned with the target listener facial reaction, which requires these diffusion models to traverse a complex denoising trajectory from these initialisations, and subsequently creates substantial opportunities for potential deviations away from the range of trajectories leading to appropriate AFRs.
Given the inherent mimicry between the human listener's and speaker's facial behaviours, we address the above denoising trajectory issue by leveraging this strong prior.
Specifically, we propose ResDiffFRG, a novel diffusion-based MAFRG framework that explicitly anchors the diffusion process to the speaker behaviour by defining its diffusion target as the residual between the speaker anchor and an AFR.
Since the speaker anchor already captures substantial target-relevant facial dynamics, the denoiser only needs to model the comparatively small, reaction-specific residual needed to transform this anchor into an AFR, rather than reconstructing the complete reaction from an unstructured state.
This requires a simpler denoising trajectory, giving fewer opportunities for potential inappropriate deviations along the trajectory, yielding a simpler generation problem.
Extensive experiments on the MARS dataset show that ResDiffFRG achieves large improvements in correlation-based appropriateness over existing methods. We conducted a denoising trajectory analysis, which showed that even at the \textbf{start of the denoising trajectory}, ResDiffFRG already achieves a higher facial-reaction correlation score than the Gaussian Diffusion baseline does after completing $60\%$ of its denoising trajectory. The full training and evaluation code will be released upon acceptance.

\end{abstract}

\section{Introduction}

During dyadic human-human conversations, the non-verbal facial behaviour that a human listener expresses in response to the given speaker's multi-modal behaviour can be treated as an appropriate facial reaction (AFR).
Such facial reactions continuously convey important social cues and emotional states. \citep{vinciarelli2009social}. 
As a result, appropriate facial reaction generation is essential for enabling virtual agents and digital humans to respond naturally to perceived multimodal human behaviours with realistic and human-like AFRs. This capability is important for creating more engaging and credible human-agent interactions 
\citep{wang2021examining}, 
with potential applications including virtual customer service assistants, realistic gaming avatars and interactive educational tutoring systems.

Early facial reaction generation (FRG) methods \citep{nojavanasghari2018interactive,huang2018generative,huang2018generating,woo2021creating,song2022learning} formulated FRG as a `one-to-one mapping' problem. They were developed to reproduce a single corresponding real AFR from the input speaker behaviour. However, human facial reactions are not fundamentally deterministic, i.e., a single speaker behaviour can elicit multiple distinct but equally appropriate AFRs, both across different listeners and from the same listener in different contexts \citep{de2012analyzing,song2023multiple}. Consequently, recent MAFRG solutions have been widely developed to explicitly model this one-to-many mapping nature through various generative architectures. These include VAE-based architectures \citep{hoque2023beamer,nguyen2024multiple}, reversible graph neural networks \citep{xu2026reversible}, adversarial frameworks (e.g. GANs) \citep{zhu2025perreactor}, and diffusion-based methods \citep{yu2023leveraging,zhu2024perfrdiff,li2025reactdiff,huang2025multiple,mao2025scattering}.
Among such methods, diffusion-based solutions are particularly effective for MAFRG. This is evidenced by the strong appropriateness and diversity of their generated AFRs as they can learn the full conditional distribution of real listeners' AFRs triggered by speaker behaviour without regressing to a single mean response. Therefore, sampling from different random noise initialisations produces multiple diverse AFRs from the same speaker behaviour, capturing the inherent one-to-many nature of MAFRG. 
Furthermore, their iterative generation process progressively constructs an AFR through a sequence of stochastic refinements, rather than requiring the full mapping from speaker to distribution to AFR to be completed in a single generation step.

\begin{wrapfigure}{r}{0.67\textwidth}
    \centering
    \includegraphics[width=0.67\textwidth]{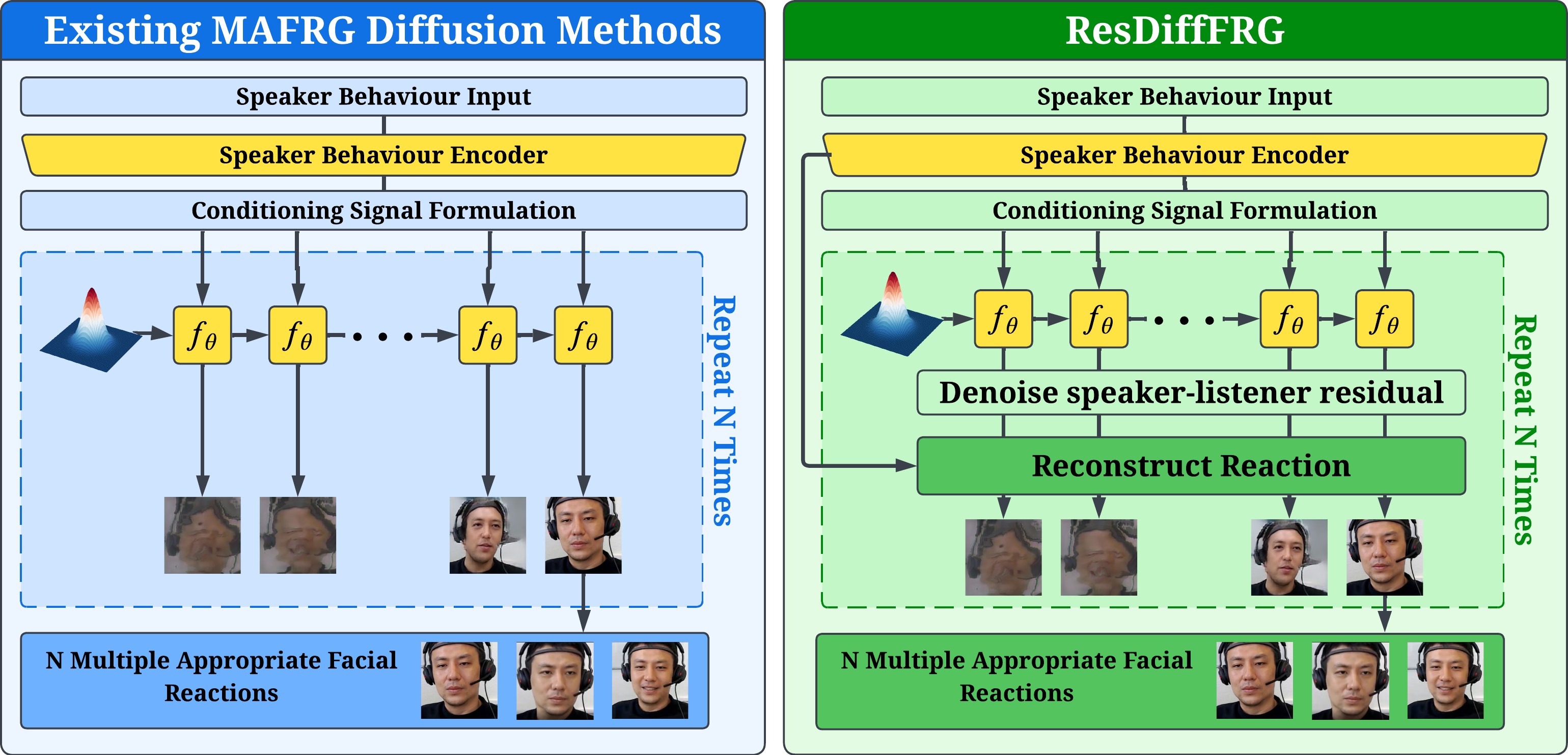}
    \caption{ResDiffFRG compared with existing methods. Left: Existing MAFRG diffusion methods attempt to denoise the full listener reaction directly from random noise, forcing the denoiser to learn complex trajectories. Right: ResDiffFRG anchors the diffusion process on the speaker anchor and learns the speaker-listener residual.}
    \label{fig:comparison}
\end{wrapfigure}

As shown in Figure \ref{fig:comparison}, such existing diffusion-based MAFRG models share a common generation mechanism that iteratively denoises random Gaussian initialisations \textbf{directly} into complete AFRs.
However, \cite{song2025reactchallenge} showed that randomly sampled Gaussian noise has extremely low facial-reaction correlation with the target real ground-truth AFRs. 
Consequently, existing diffusion-based MAFRG models begin from an initial state that is inherently poorly aligned with the desired facial reaction.
The denoiser must therefore learn a relatively large transformation from an uninformative random initialisation to a \textbf{complete} appropriate listener response.
\textbf{This requires complex denoising trajectories, which have substantial opportunities for potential deviations away from the range of trajectories leading to appropriate AFRs.}


Importantly, the mimicry between speakers and listeners in dyadic interactions provides a natural structural prior for FRG, which existing diffusion methods do not exploit. Physiological studies have long established that observing another person's facial expression can induce rapid, expression-congruent facial muscle responses \citep{dimberg1982facial}. More recently, 
\cite{song2025reactchallenge} showed that simply copying the speaker's facial behaviour provides a strong baseline for MAFRG, yielding a facial reaction correlation score that is 1633\% higher than that of random Gaussian noise. 
These findings suggest that, unlike other generative tasks, diffusion-based MAFRG models do not need to begin from an entirely uninformative initial state of random Gaussian noise, given that a substantially more target-relevant estimate (the speaker's own behaviour) is already available.

Motivated by this observation, we propose a novel \textbf{Residual Diffusion for Facial Reaction Generation (ResDiffFRG) framework} for the offline MAFRG task. The central departure from existing diffusion-based approaches is that ResDiffFRG does \textbf{not} directly diffuse and reconstruct the complete listener AFR. Instead, it explicitly anchors the diffusion process on the speaker's facial behaviour and defines the diffusion target as the \textbf{residual (difference) between the listener's and speaker's facial behaviours}. 
\textbf{This reformulation changes both the role of the speaker information and the learning objective of the diffusion model}. In existing approaches, speaker behaviour is typically treated as contextual or conditioning information while the model still generates the listener AFR in its entirety. In ResDiffFRG, by contrast, the speaker trajectory becomes an explicit reference point in the generation space. Because this anchor already captures part of the temporal and behavioural structure of the target AFR, the denoiser only needs to model the comparatively smaller, listener-specific deviations required to transform the speaker behaviour into an appropriate reaction. This reduces the magnitude and complexity of the transformation required throughout the reverse diffusion process and yields a more constrained, task-specific denoising trajectory.
The main contributions and novelties of this paper are summarised as follows:
\begin{itemize}

    \item \textbf{We reformulate facial reaction generation from direct AFR generation into residual generation between listener and speaker behaviours.} Unlike existing diffusion-based MAFRG approaches, where the diffusion target is the complete listener's AFR, ResDiffFRG anchors the diffusion process on the speaker behaviour rather than using the speaker merely as conditioning information. \textbf{To the best of our knowledge, this is the first diffusion-based MAFRG framework to formulate the task in this way.}

    \item \textbf{We further introduce dimension-wise residual normalisation using statistics estimated from the training set.} This addresses scale imbalance across residual dimensions, produces a better-conditioned diffusion target, and \textbf{substantially improves temporal alignment compared with directly diffusing unnormalised residuals.} 

    \item Extensive experiments show that \textbf{both ResDiffFRG (direct) and ResDiffFRG (normalised) offer large improvements in correlation-based appropriateness (FRCorr) over existing methods.} 
    We conducted a denoising-trajectory analysis, which further shows that residual diffusion is appropriate from the first denoising step.
    
\end{itemize}


\section{Related Work}

Early facial reaction generation (FRG) models treated FRG as a one-to-one mapping from speaker behaviour to a single ground-truth listener reaction. An early example is DyadGAN \citep{huang2017dyadgan}, which adopted a two-stage conditional Generative Adversarial Network (GAN) to predict listener facial reactions from the speaker's facial action units. Subsequent one-to-one FRG approaches include \cite{nojavanasghari2018interactive, zhao2018personality, team2021creating, shao2021personality, ng2022learning, song2022learning, woo2023amii}. However, FRG is an inherently non-deterministic task \citep{song2023multiple}, and so these one-to-one mapping models cannot capture the diversity of the range of AFRs that are appropriate to a single speaker stimulus.

More recently, the field of multiple appropriate facial reaction generation (MAFRG) have been developed to address the non-deterministic nature of facial reactions. FRDiff \citep{yu2023leveraging} was the first work to introduce diffusion into MAFRG. Following this work, further diffusion-based MAFRG approaches have been proposed, including VQ-Diff \citep{nguyen2024vector} which operates diffusion over vector-quantised discrete latent codes of facial reactions, ReactDiff \citep{luo2025reactdiff} which incorporates temporal facial behavioural kinematic constraints and facial AU dependency constraints into the diffusion process, 
SC-Diff \citep{mao2025scattering} which provides structured temporal information via scattering representations to the denoising process, \cite{huang2025multiple} incorporated multi-view transformations of the speaker video in the denoising process, and SAFRG \citep{liu2026safrg} which aligned the diffusion process with the target listener speech. Other recent diffusion-based MAFRG methods have explicitly focused on personalisation. These include PerFRDiff \citep{zhu2024perfrdiff} which incorporates personalised weight editing into the diffusion process, and PerReactor \citep{zhu2025perreactor} which uses a hierarchical decoupling-fusion framework to generate the conditioning signal used in the denoising process. 

However, these diffusion models all share a common denominator: they all seek to transform some random noise representation directly into an AFR's representation. They only differ in the conditioning signal used to guide this transformation process. Given the poor alignment between random noise and real AFRs, these models are required to learn a difficult denoising trajectory. Given the inherent mimicry between the listener's and speaker's facial behaviours in dyadic conversations, our ResDiffFRG framework avoids the need to learn such difficult denoising trajectories by anchoring the denoising process on the speaker's own facial behaviour representation.

\section{Problem Formulation}

The offline MAFRG task aims to learn a model $H$ that takes in the given full speaker behaviour sequence $b_S^{1:T}$ and accordingly generates multiple AFR sequences $P_L(b_S^{1:T}) = \{p_L(b_S^{1:T})_1, \dots, p_L(b_S^{1:T})_N\}$. This task requires that each generated AFR $p_L(b_S^{1:T})_i \in P_L(b_S^{1:T})$ is similar to at least one provided real AFR $f_L(b_S^{1:T})_j \in F_L(b_S^{1:T})$ expressed by human listeners \citep{song2023multiple} as:
\begin{equation}
\begin{aligned}
    P_L(b_S^{1:T}) &= H(b_S^{1:T}), \\
    &\quad \forall i \in \{1,\ldots,N\},\ 
    \exists j \in \{1,\ldots,M\},\ 
    \text{s.t.}\ 
    p_L(b_S^{1:T})_i \sim f_L(b_S^{1:T})_j .
\end{aligned}
\end{equation}
where $\sim$ denotes the similarity between two AFR sequences.

\begin{figure*}[!t]
  \centering
  \includegraphics[width=0.95\textwidth]{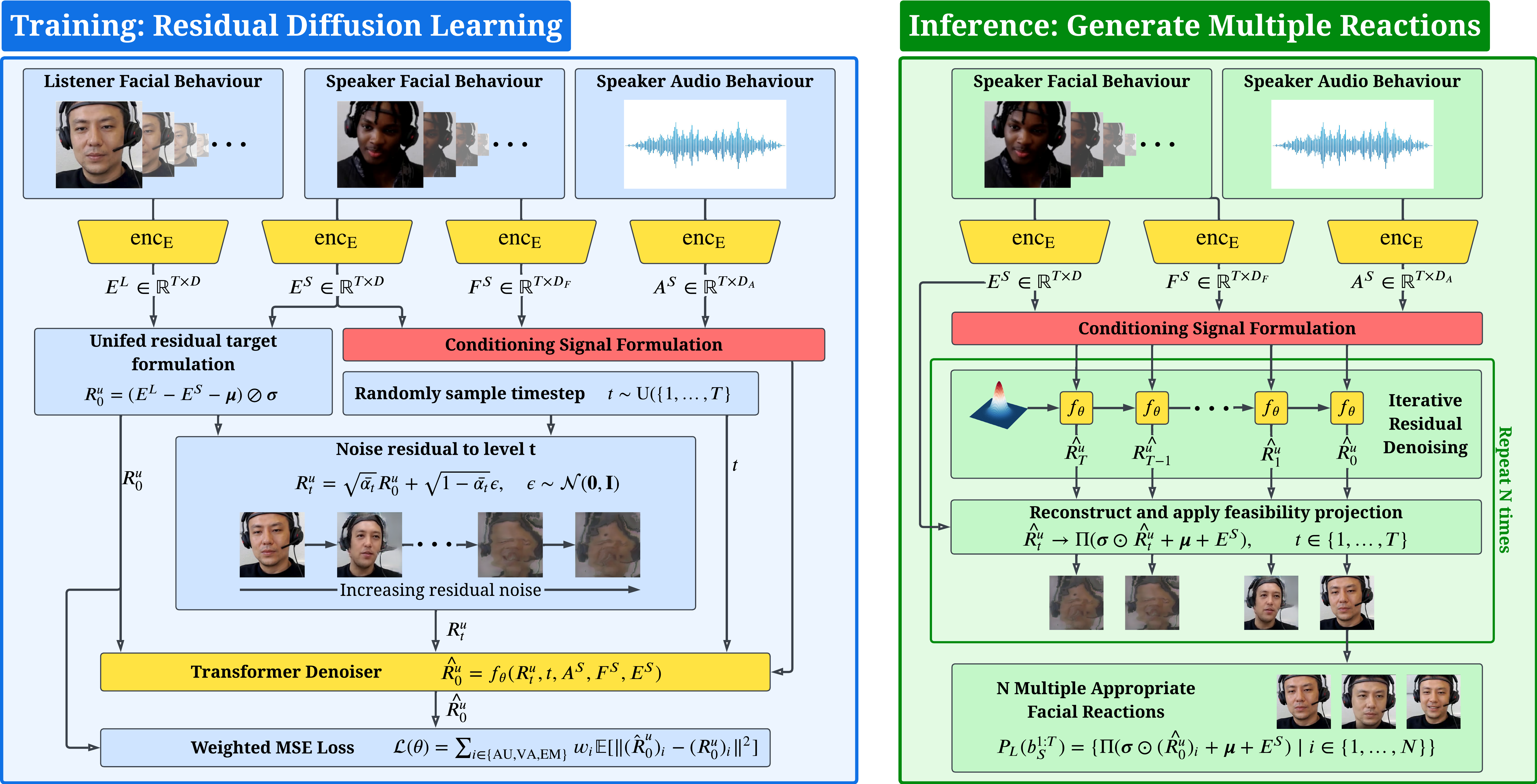}
  \caption{ResDiffFRG Framework Pipeline. Left: During training, the residual diffusion target is jointly constructed from the encoded listener and speaker features, noised using the forward diffusion process and the residual denoiser is trained to recover the clean residual. Right: During inference, DDIM is used to sample N residuals, which are reconstructed back into the listener behaviour space using the speaker anchor and passed through a feasibility projection.}
\label{fig:pipeline}
\end{figure*}

\section{Methodology}



\subsection{Overview}

The proposed ResDiffFRG framework is illustrated in Figure \ref{fig:pipeline}. Given an input speaker behaviour sequence $b_S^{1:T}$, multimodal behavioural representations are extracted using pre-trained feature encoders. Specifically, $D_A$-dimensional audio representations are extracted using encoder $\mathrm{enc_A}$ \citep{baevski2020wav2vec}, $D_F$-dimensional facial behaviour representations using encoder $\mathrm{enc_F}$ \citep{wang2022faceverse} and $D$-dimensional emotional embeddings using encoder $\mathrm{enc_E}$ \citep{luo2022learning}:
\begin{equation}
    A^S = \mathrm{enc_A}(b_S^{1:T}) \in \mathbb{R}^{T \times D_A}, \quad
    F^S = \mathrm{enc_F}(b_S^{1:T}) \in \mathbb{R}^{T \times D_F}, \quad
    E^S = \mathrm{enc_E}(b_S^{1:T}) \in \mathbb{R}^{T \times D}
\end{equation}
where $A^S, F^S, E^S$ denote the encoded audio, facial and emotion behaviours expressed by the speaker, respectively.
Similarly, each paired real ground-truth listener AFR $f_L^{1:T}$ is encoded using $\mathrm{enc_E}$ to obtain the target listener AFR representation $E^L$, which is used during training, as follows:

\begin{equation}
    E^L = \mathrm{enc_E}(f_L^{1:T}) \in \mathbb{R}^{T \times D}
\end{equation}

Unlike existing MAFRG diffusion models that generate the complete AFR representation directly from the sampled Gaussian noise, our ResDiffFRG introduces a \textbf{speaker-anchored residual} that anchors the diffusion process on $E^S$ and so the model is only required to generate the AFR-specific deviation from $E^S$. This deviation can be expressed as our residual target $R_0$, which is defined as:
\begin{equation}
    R_0 = E^L - E^S \in \mathbb{R}^{T \times D}
    \label{eq:residual_def}
\end{equation}
By exploiting the speaker-listener facial behaviour mimicry, ResDiffFRG anchors the diffusion process on $E^S$, only requiring $R_0$ to be modelled, establishing a more target-relevant starting point for generation. Unlike existing MAFRG diffusion models, ResDiffFRG avoids the need to traverse complex denoising trajectories to reconstruct the entire listener behaviour $E^L$ from random noise.

To account for substantially different variances across residual dimensions, we further introduce a \textbf{normalised residual formulation} (Section \ref{subsec:normalised_residual}). Together with the direct residual formulation (as defined in Equation \ref{eq:residual_def}), both formulations can be represented through the unified residual target $R_0^u$:
\begin{equation}
  R_0^u = (E^L - E^S - \boldsymbol{\mu}) \oslash \boldsymbol{\sigma}
  \label{eq:overview_unified_residual}
\end{equation}
where $\oslash$ denotes element-wise division. $(\boldsymbol{\mu}, \boldsymbol{\sigma}) = (\mathbf{0}, \mathbf{1})$ for the direct formulation, while the normalised formulation uses per-dimension residual means and standard deviations calculated from the training set.

\textbf{Training strategy:} A standard forward diffusion process \citep{ho2020denoising} is applied to the unified residual target $R^u_0$, as defined in Eq \ref{eq:overview_unified_residual}:

\begin{equation}
    R^u_t = \sqrt{\bar{\alpha_t}} R^u_0 + \sqrt{1 - \bar{\alpha_t}}\epsilon, \qquad \epsilon \sim \mathcal{N}(\mathbf{0}, \mathbf{I})
    \label{eq:unified_forward_diffusion}
\end{equation}

Subsequently, we train our \textbf{residual denoiser} $f_\theta$ (Section \ref{subsec:residual_denoiser}) to estimate the clean unified residual target $R^u_0$ by predicting $\hat{R}^u_0$, conditioned on $A^S, F^S, E^S$:

\begin{equation}
    \hat{R}^u_0 = f_\theta(R_t^u, t, A^S, F^S, E^S)
    \label{eq:denoiser_definition}
\end{equation}

This formulation enables $f_\theta$ to learn the \textbf{conditional residual distribution} $p_\theta(R_0^u|A^S, F^S, E^S)$, rather than the more complex direct conditional distribution $p_\theta(E^L|A^S, F^S, E^S)$ learned by existing MAFRG diffusion methods. Furthermore, to enable classifier-free guidance (CFG), the conditioning signals $A^S, F^S, E^S$ are jointly dropped with probability $p_{\mathrm{drop}}$ to allow $f_\theta$ to learn both conditional and unconditional distributions.

\textbf{Generation Multiple AFRs during Inference:} During inference, DDIM \citep{song2020denoising} with CFG is employed to sample $N$ unified residuals from the learned conditional residual distribution $p_\theta(R_0^u|A^S, F^S, E^S)$. 
At each denoising step, we generate a prediction $\hat{R^u_0}$ for the clean residual target, while applying CFG with weight $w$:

\begin{equation}
    \hat{R^u_0} = f_\theta(R^u_t, t, \varnothing, \varnothing, \varnothing) + w(f_\theta(R^u_t, t, A^S, F^S, E^S) - f_\theta(R^u_t, t, \varnothing, \varnothing, \varnothing))
\end{equation}

From $\hat{R^u_0}$, the noise estimate required by the DDIM update is obtained by inverting Eq. \ref{eq:unified_forward_diffusion} for $\epsilon$. This sampling procedure is repeated $N$ times from $N$ independent initial noise samples to obtain a set of $N$ unified residuals $\{(\hat{R}_0^u)_1, \dots, (\hat{R}_0^u)_N\}$. Each residual is inverse normalised (for the normalised variant) and then added to the speaker anchor. Finally, a fixed feasibility transformation $\Pi$ is applied to each reconstructed sequence to ensure that the mixed-type structure of our target listener emotion signal is respected. Dimensions that take binary values are rounded to the nearest $\{0, 1\}$, while dimensions that take continuous values are clipped to their 1st and 99th percentiles of all listener reactions in the training split (\textbf{no} validation or test data are used). The final $N$ generated AFRs are formulated as:
\begin{equation}
    P_L(b_S^{1:T}) = \{\Pi(\boldsymbol{\sigma} \odot (\hat{R_0^u})_i + \boldsymbol{\mu} + E^S) \; | \; i \in \{1, \dots, N\}\}
    \label{eq:final_projection}
\end{equation}

where $\odot$ denotes element-wise multiplication.

\subsection{Normalised Residual Formulation}
\label{subsec:normalised_residual}

Although Equation \ref{eq:residual_def} provides a speaker-anchored diffusion target $R_0 \in \mathbb{R}^{T \times D}$, \textbf{the 
$D$ dimensions of $R_0$ have substantially different variances}, as the $D$ channels of the facial-attribute signal have varying channel types (e.g., there are binary action units but also continuous valence/arousal values). 
Consider $j$-th residual dimension of $R_0$, $(R_0)_j=E_j^L - E_j^S$. Let $\sigma_j$ denote its standard deviation. The standard forward diffusion process \citep{ho2020denoising} is formulated as:
\begin{equation}
    (R_t)_j = \sqrt{\bar{\alpha_t}} (R_0)_j + \sqrt{1 - \bar{\alpha_t}}\epsilon
    \label{eq:dimension_forward_diffusion}
\end{equation}
where the signal term has variance $\bar{\alpha_t} \sigma_j^2$ and the noise term has variance $1 - \bar{\alpha_t}$. The per-dimension signal-to-noise ratio (SNR) is therefore represented as: 
\begin{equation}
    \mathrm{SNR}_{t, j} = \frac{\bar{\alpha_t} \sigma_j^2}{1 - \bar{\alpha_t}}
    \label{eq:snr}
\end{equation}
As a result, a dimension's SNR is dependent on its standard deviation. 
Since different residual dimensions have varying standard deviations, a given timestep $t$ can correspond to different relative corruption levels across dimensions. 
Consequently, the same diffusion timestep does not have a consistent meaning across residual dimensions, as some dimensions are substantially more corrupted than others. 

To address this inconsistency, our ResDiffFRG (normalised) standardises each residual dimension using statistics calculated exclusively from the training set. We define the statistics $\bar{\mu}, \bar{\sigma}$ and the normalised residual target $\tilde{R}$ for each dimension as:
\begin{equation}
    \bar{\mu_j} = \mathbb{E}_{\mathrm{train}}[(R_0)_j], \qquad \bar{\sigma_j} = \sqrt{\mathrm{Var}_{\mathrm{train}}[(R_0)_j]}, \qquad (\tilde{R}_0)_j = \frac{(R_0)_j - \bar{\mu_j}}{\bar{\sigma_j}}
\end{equation}
This also ensures that the normalised residual target is consistent with the unified residual target $R^u_0$ from Eq. \ref{eq:overview_unified_residual}, with $\boldsymbol{\mu} =\bar{\mu}$ and $\boldsymbol{\sigma} = \bar{\sigma}$.

Assuming that for any dimension $j$, $\bar{\sigma_j} \approx \sigma_j$, we get that $\mathrm{Var}[\tilde{R}_{0, j}] = \frac{\mathrm{Var}[R_{0, j}]}{{\sigma_j}^2} = \frac{{\bar{\sigma_j}}^2}{{\sigma_j}^2} \approx 1$.
Applying the same forward diffusion process to $\tilde{R}_0$ yields:
\begin{equation}
    \tilde{R_t} = \sqrt{\bar{\alpha_t}} \tilde{R_0} + \sqrt{1 - \bar{\alpha_t}}\epsilon, \qquad \epsilon \sim \mathcal{N}(\mathbf{0}, \mathbf{I})
    \label{eq:normalised_diffusion}
\end{equation}

Consequently, the normalised formulation makes the SNR of each dimension at each timestep independent of the original scale of each residual dimension (i.e., its standard deviation) as:
\begin{equation}
    \mathrm{SNR}_{t, j}^\mathrm{norm} = \frac{\bar{\alpha_t} \mathrm{Var}[\tilde{R}_{0,j}]}{1 - \bar{\alpha_t}} \approx \frac{\bar{\alpha_t}}{1 - \bar{\alpha_t}}
    \label{eq:snr_normalised}
\end{equation}
This ensures that all diffusion timesteps carry a consistent meaning across all residual dimensions.

\subsection{Residual Denoiser}
\label{subsec:residual_denoiser}


Our residual denoiser $f_\theta$, as defined in Eq \ref{eq:denoiser_definition}, is implemented using a transformer decoder. The initial noisy $R_t^u$ is linearly projected to form the query stream. The encoded speaker representations $A^S, F^S$ and $E^S$ are linearly projected and concatenated along the time dimension together with a sinusoidal timestep embedding to form a conditioning memory. Each layer attends to this conditioning memory via cross-attention.

The $D$-dimensional output consists of $D_\mathrm{AU}$ binary action channels, $D_\mathrm{VA}$ valence/arousal channels and $D_\mathrm{EM}$ emotion-probability channels. Let $(\hat{R}_0^u)_\mathrm{AU}, (\hat{R}_0^u)_\mathrm{VA}, (\hat{R}_0^u)_\mathrm{EM}$ denote the corresponding partitions of $\hat{R}_0^u$ into action units, valence/arousal values and emotion probabilities, respectively. $f_\theta$ is trained using a weighted mean squared error (MSE) objective:
\begin{equation}
    \mathcal{L}(\theta) = \sum_{i \in \{\mathrm{AU}, \mathrm{VA}, \mathrm{EM}\}} w_i \mathbb{E}[\|(\hat{R}_0^u)_i - (R_0^u)_i\|^2]
\end{equation}


\label{subsec:condition_denoise}





\section{Experiments}

\subsection{Experimental settings}

\textbf{Datasets:} All ResDiffFRG variants used in the ablation experiments were trained, validated and tested on the Multi-modal Multiple Appropriate Reaction in Social Dyads (MARS) dataset using the official train/validation/test splits defined by the REACT 2025 \citep{song2025reactchallenge}. This dataset contains dyadic sessions with multiple speaker-listener recordings captured under the same setting, allowing listener clips within each session to form a set of AFRs for evaluation. In total, the dataset comprises 2,856 dyadic interactions, with 1,660 allocated to training, 571 to validation and 625 to testing.

\textbf{Implementation details:} Both ResDiffFRG variants and all models used in ablation experiments employed the denoiser architecture outlined in Section \ref{subsec:residual_denoiser}, comprising 9 layers, 8 heads and 34,270,233 trainable parameters. All models were trained for 500 epochs using the AdamW optimizer with a learning rate of $10^{-4}$ and a batch size of 8. Validation was performed every 10 epochs, and the checkpoint with the lowest validation concordance correlation loss was selected to ensure comparability between variants. Following PerFRDiff \citep{zhu2024perfrdiff}, all models except the ResDiffFRG (normalised) model used $w_{AU} = 1, w_{VA}=5, w_{EM} = 5$. For ResDiffFRG (normalised), since normalisation ensures that the SNR is approximately equal across dimensions, we removed the variance-adjusting weights introduced by PerFRDiff and gave the three attribute groups equal weighting by setting $w_{AU} = w_{VA} = w_{EM} = 1$. A minimal inference demo, including a lightweight model checkpoint and example input, is provided in the supplementary material. 

\textbf{Metrics:} We adopted the standard MAFRG metrics (FRCorr, FRDist, FRDiv, FRVar and FRSyn) in evaluation as introduced by \citep{song2023multiple}. Further detailed descriptions are provided in Appendix \ref{appendix:metrics}.

\subsection{Comparison with existing solutions}

\begin{table*}[!t]
\caption{Comparison with existing methods}
\label{table:comparison}
\centering

\begin{tabular}{lccccc}

\toprule
\textbf{Method} & \multicolumn{2}{c}{\textbf{Appropriateness}} & \multicolumn{2}{c}{\textbf{Diversity}} & \textbf{Synchrony} \\

\cmidrule(lr){2-3}
\cmidrule(lr){4-5}
\cmidrule(lr){6-6}

& FRCorr ($\uparrow$) & FRDist ($\downarrow$) & FRDiv ($\uparrow$) & FRVar ($\uparrow$) & FRSyn ($\downarrow$) \\

\midrule

GT & 10.00 & 0.00 &  0.1876 & 0.0669 & 48.66\\
B\_Random & 0.03 & 474.68 & 0.3342 & 0.1671 & 46.64\\
B\_Mime & 0.52 & 206.02 & 0.0000 & 0.0766 & 43.70\\
B\_MeanFR & 0.00 & 205.65 & 0.0000 & 0.0000 & 49.00\\

\midrule

Trans-VAE & 0.24 & 158.97 & 0.0079 & 0.0067 & 49.00\\
REGNN & 0.46 & 141.93 & 0.0010 & 0.0036 & 45.52\\
PerFRDiff & 0.56 & 177.76 & 0.1386 & 0.0706 & 48.54\\

\midrule

ResDiffFRG (normalised) & 0.78 & 180.70 & 0.0968 & 0.0725 & 45.73 \\
ResDiffFRG (direct) & 0.80 & 198.96 & 0.1461 & 0.0871 & 47.94\\

\bottomrule

\end{tabular}

\end{table*}

As shown in Table \ref{table:comparison}, both the direct and normalised variants of our proposed ResDiffFRG framework outperform the best REACT 2025 baseline model (PerFRDiff) in terms of FRCorr by around 43\% and 39\% respectively, with even greater improvements when compared with Trans-VAE and REGNN. Trans-VAE and REGNN attain low FRDiv (0.0079 and 0.0010, respectively) and low FRVar (0.0067 and 0.0036, respectively). Therefore, despite attaining a better FRDist score than PerFRDiff and both ResDiffFRG variants, their low FRDiv and low FRVar scores suggest that their FRG process is almost deterministic, falling short of the one-to-many requirement of the MAFRG task. PerFRDiff has the best FRDist score among the models that exhibit genuine diversity, and the normalised ResDiffFRG variant's FRDist score is within 1.7\% of that of PerFRDiff.

\subsection{Ablation studies}

\begin{table*}[!t]
\caption{Effect of Normalised Residual Diffusion over other diffusion targets and Effect of CFG}
\label{table:ablation}
\centering

\begin{tabular}{lccccc}

\toprule
\textbf{Method} & \multicolumn{2}{c}{\textbf{Appropriateness}} & \multicolumn{2}{c}{\textbf{Diversity}} & \textbf{Synchrony} \\

\cmidrule(lr){2-3}
\cmidrule(lr){4-5}
\cmidrule(lr){6-6}

& FRCorr ($\uparrow$) & FRDist ($\downarrow$) & FRDiv ($\uparrow$) & FRVar ($\uparrow$) & FRSyn ($\downarrow$) \\

\midrule

Direct (Gaussian) Diffusion w/o CFG & 0.64 & 195.02 & 0.1371 & 0.0643 & 48.56\\
Direct (Gaussian) Diffusion w CFG & 0.69 & 195.18 & 0.1450 & 0.0717 & 48.47\\
ResDiffFRG (direct) w/o CFG & 0.78 & 199.84 & 0.1474 & 0.0864 & 46.19 \\
ResDiffFRG (direct) & 0.80 & 198.96 & 0.1461 & 0.0871 & 47.94\\
ResDiffFRG (normalised) w/o CFG & 0.74 & 183.73 & 0.0857 & 0.0740 & 45.73 \\
ResDiffFRG (normalised) & 0.78 & 180.70 & 0.0968 & 0.0725 & 45.73 \\

\bottomrule

\end{tabular}

\end{table*}

\begin{table*}[!t]
\caption{Effect of Diffusion Transformer}
\label{table:ablation_transformer}
\centering

\begin{tabular}{lccccc}

\toprule
\textbf{Method} & \multicolumn{2}{c}{\textbf{Appropriateness}} & \multicolumn{2}{c}{\textbf{Diversity}} & \textbf{Synchrony} \\

\cmidrule(lr){2-3}
\cmidrule(lr){4-5}
\cmidrule(lr){6-6}

& FRCorr ($\uparrow$) & FRDist ($\downarrow$) & FRDiv ($\uparrow$) & FRVar ($\uparrow$) & FRSyn ($\downarrow$) \\

\midrule

Stochastic Baseline & 0.43 & 333.52 & 0.1668 & 0.1170 & 45.73 \\
ResDiffFRG (normalised) & 0.78 & 180.70 & 0.0968 & 0.0725 & 45.73 \\

\bottomrule

\end{tabular}

\end{table*}

In Table \ref{table:ablation}, all three models employ the \textbf{same} transformer denoiser architecture with the \textbf{same} parameter count, noise schedule, train/val/test splits and conditioning signals. The primary distinction lies in their diffusion target. Specifically, the diffusion target for the direct (Gaussian) diffusion model (rows 1-2) is the listener reaction $E^L$, whereas the diffusion targets for the two variants of ResDiffFRG (rows 3-6) are the direct and normalised speaker-listener residuals, as described in the Methodology section. 

Table \ref{table:ablation} demonstrates that changing the diffusion target from the listener reaction $E^L$ to the speaker-listener residual raises the FRCorr, both with and without CFG, while all other experimental settings remain unchanged. This improvement cannot be attributed to the control factors outlined in the previous paragraph (such as denoiser architecture and conditioning signals), as they were kept constant. Similarly, the improvement could not be due to a difference in training losses, as both the Direct Gaussian Diffusion and ResDiffFRG (direct) employ the identical training and validation losses. The gain in FRCorr is not achieved at the expense of diversity, as FRDiv and FRVar between the Direct Gaussian Diffusion model and the ResDiffFRG (direct) model are very close. These findings indicate that the choice of the speaker-listener residual as the diffusion target can be isolated as the main source of improvement in FRCorr between the Direct Gaussian Diffusion ablation model and ResDiffFRG (direct). 

Furthermore, Table \ref{table:ablation} shows that normalisation improves FRDist for the ResDiffFRG, as the normalised ResDiffFRG model has the lowest FRDist amongst all ablation models, much lower than direct ResDiffFRG and direct Gaussian diffusion. As a result, despite a small cost on FRCorr, by standardising SNR across all dimensions (for each timestep) through normalisation and a uniform MSE training loss, the diffusion target becomes better-conditioned. Additionally, Table \ref{table:ablation} shows that CFG improves model performance for all three ablation models, and Table \ref{table:ablation_transformer} further demonstrates that simply sampling from the residual distribution and adding it to the speaker anchor (row 1) achieves around half of the FRCorr as ResDiffFRG and produces a much worse FRDist, which highlights the important role of our denoiser architecture. 

\section{Denoising Trajectory Analysis}

\begin{table}[!t]
\caption{Appropriateness (FRCorr) During Denoise Process}
\label{table:denoise_trajectory}
\centering
\begin{tabular}{lcccccc}
\toprule
\textbf{Denoise Fraction} & 0.0 & 0.2 & 0.4 & 0.6 & 0.8 & 1.0 \\
\midrule
Direct Gaussian & 0.06 & 0.13 & 0.23 & 0.39 & 0.62 & 0.69 \\
ResDiffFRG & 0.43 & 0.50 & 0.61 & 0.72 & 0.77 & 0.78 \\
\bottomrule
\end{tabular}
\end{table}

\begin{figure*}[!t]
  \centering
  \includegraphics[width=\textwidth]{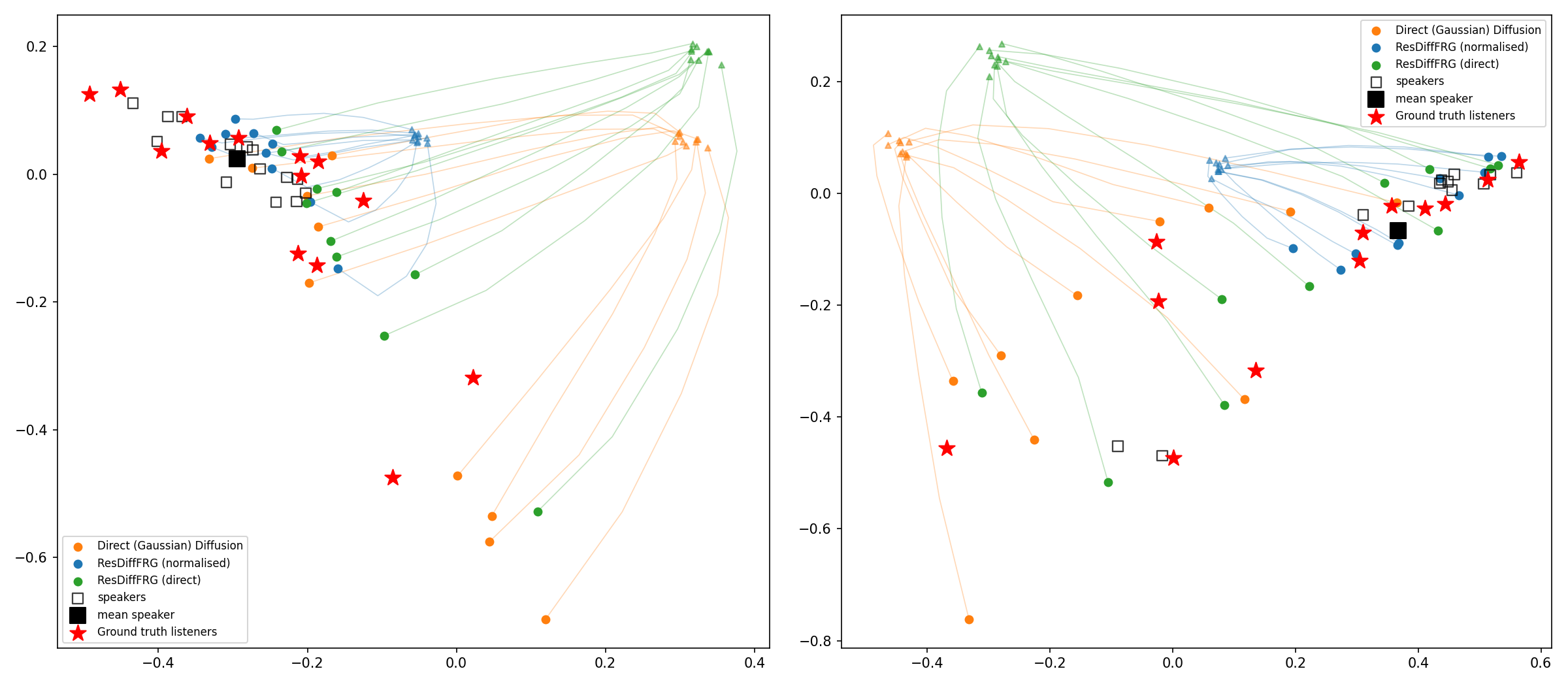}
  \caption{Denoising trajectories in mean-emotion PCA space. Two different sessions are presented. For each model, the mean speaker behaviour for each session is used as conditioning and their DDIM iterates are tracked and traced during the denoise process. For each iterate, we isolate the temporal mean of their $D_{\mathrm{EM}}$-dimensional emotion channels. They are reconstructed back to listener reaction space (where required) and passed through the feasibility projection $\Pi$. The trajectories formed by these iterates are plotted using a PCA fit. The triangles mark the start points, circles mark the final generation result, red stars mark the ground truth listener sequences, white squares mark the speaker sequences and the black square marks the mean speaker sequence.}
\label{fig:resdiff_pca}
\end{figure*}

To further investigate the source of the gain in Table \ref{table:ablation}, for both ResDiffFRG (normalised) and Direct Gaussian Diffusion, their DDIM iterates were tracked and traced every 10 sampling steps during their 50-step DDIM process. For ResDiffFRG, each iterate $R^u_t$ was reconstructed back to listener reaction space, $\boldsymbol{\sigma} \odot R^u_t+\boldsymbol{\mu}+E^S$. For the Direct Gaussian model, its iterates were already in listener reaction space and therefore required no reconstruction. Then, for both models, their iterates were passed through the feasibility projection $\Pi$ to ensure commensurability. For both models, 10 sequences of iterates were sampled and processed as described. They were used to calculate the FRCorr at denoise fractions $0.0, 0.2, \dots,1.0$ (every 10 sampling steps) presented in Table \ref{table:denoise_trajectory}.

The results in Table \ref{table:denoise_trajectory} provide empirical support for our hypothesis that ResDiffFRG simplifies the denoising trajectory. At the beginning of the denoising trajectory, ResDiffFRG achieves an FRCorr of 0.43, which is not achieved by the Direct Gaussian baseline until more than 60\% of its denoising trajectory has been completed. This suggests that the speaker anchor already provides the coarse speaker-listener correspondence that the Direct Gaussian baseline must instead recover through denoising from an unstructured initialisation. \textbf{Therefore, ResDiffFRG requires a smaller transformation over its trajectory, as opposed to first searching for the rough appropriate region of listener-reaction space.} Instead, it can devote its denoising steps to modelling and refining the reaction-specific deviation from the speaker anchor. Figure \ref{fig:resdiff_pca} qualitatively reflects this difference in trajectories. These results are consistent with our central hypothesis that ResDiffFRG reduces the required transformation, giving us a simpler denoising problem with reduced opportunities to deviate from the range of trajectories leading to appropriate AFRs.

\section{Conclusion}

To conclude, we propose ResDiffFRG, \textbf{the first speaker-anchored residual diffusion framework that reformulates diffusion-based MAFRG}. Instead of generating a complete AFR from random Gaussian noise, ResDiffFRG exploits speaker-listener mimicry by anchoring diffusion to speaker behaviour and modelling only the residual required to obtain an AFR.
We further introduce a normalised variant to approximately equalise all residual dimensions' SNRs at each diffusion timestep. Experiments show that both variants substantially improve FRCorr over existing methods. Controlled ablations isolate the residual diffusion target formulation as the main source of this gain. Our denoising-trajectory analysis supports our central hypothesis that speaker anchoring greatly simplifies the denoising problem, as ResDiffFRG begins with an FRCorr that the Direct Gaussian Diffusion baseline does not achieve until after 60\% of its trajectory. 
More broadly, in scenarios where there is substantial mimicry between the input signal and the target signal,
our findings suggest that input-anchored residual diffusion can provide a more effective and better-constrained alternative to full-target generation from unstructured noise, \textbf{offering a potentially generalisable direction for conditional diffusion beyond MAFRG.}

\bibliography{main}
\bibliographystyle{main}

\appendix
\section{Appendix}

\subsection{Qualitative Generation Example}

Figure 4 presents a qualitative example of reactions generated by ResDiffFRG. Given the same speaker behaviour sequence, we independently sampled and visualised three generated AFRs based on a single speaker behaviour sequence. They exhibit distinct facial expressions, illustrating the one-to-many nature of the learned reaction distribution. The generated reactions are also noticeably distinct from the speaker behaviour itself, indicating that ResDiffFRG does not simply copy the speaker anchor, but learns the reaction-specific residuals that transform the anchor into different listener reactions.

\subsection{Further Details about Evaluation Metrics}

\label{appendix:metrics}

We conducted all evaluations in this paper based on the five primary MAFRG metrics proposed in \citep{song2023multiple}. These metrics have been summarised in \citep{liu2026safrg}:
\begin{itemize}
    \item \textbf{FRCorr} is measured by Concordance Correlation Coefficient (CCC) that computes the correlation and agreement between the generated facial reaction attribute sequence and its ground-truth facial reaction attribute sequence;

    \item \textbf{FRDist} is measured by Dynamic Time Warping (DTW) that computes distances between the generated facial reaction attribute sequence and its ground-truth facial reaction attribute sequence;

    \item \textbf{FRDiv}: the facial reaction diverseness metric measures diverseness among generated facial reactions by calculating the sum of the MSE scores of each generated facial reaction pairs for each input speaker behaviour, and then returning the mean of these sums across all speaker behaviours. Higher values indicate higher diversity between the different generated facial reactions.

    \item \textbf{FRVar}: the facial reaction variance metric calculates the average variance across frames for each generated facial reaction. Higher values indicate higher diversity within each generated facial reaction.

    \item \textbf{FRSyn} is measured by time-Lagged Cross-Correlation (TLCC) that computes synchrony between the generated facial reaction attribute sequence and its ground-truth facial reaction attribute sequence, i.e. how strongly the listener facial behaviour follows the speaker facial behaviour. 
\end{itemize}

\begin{figure}
  \centering
  \includegraphics[width=\textwidth]{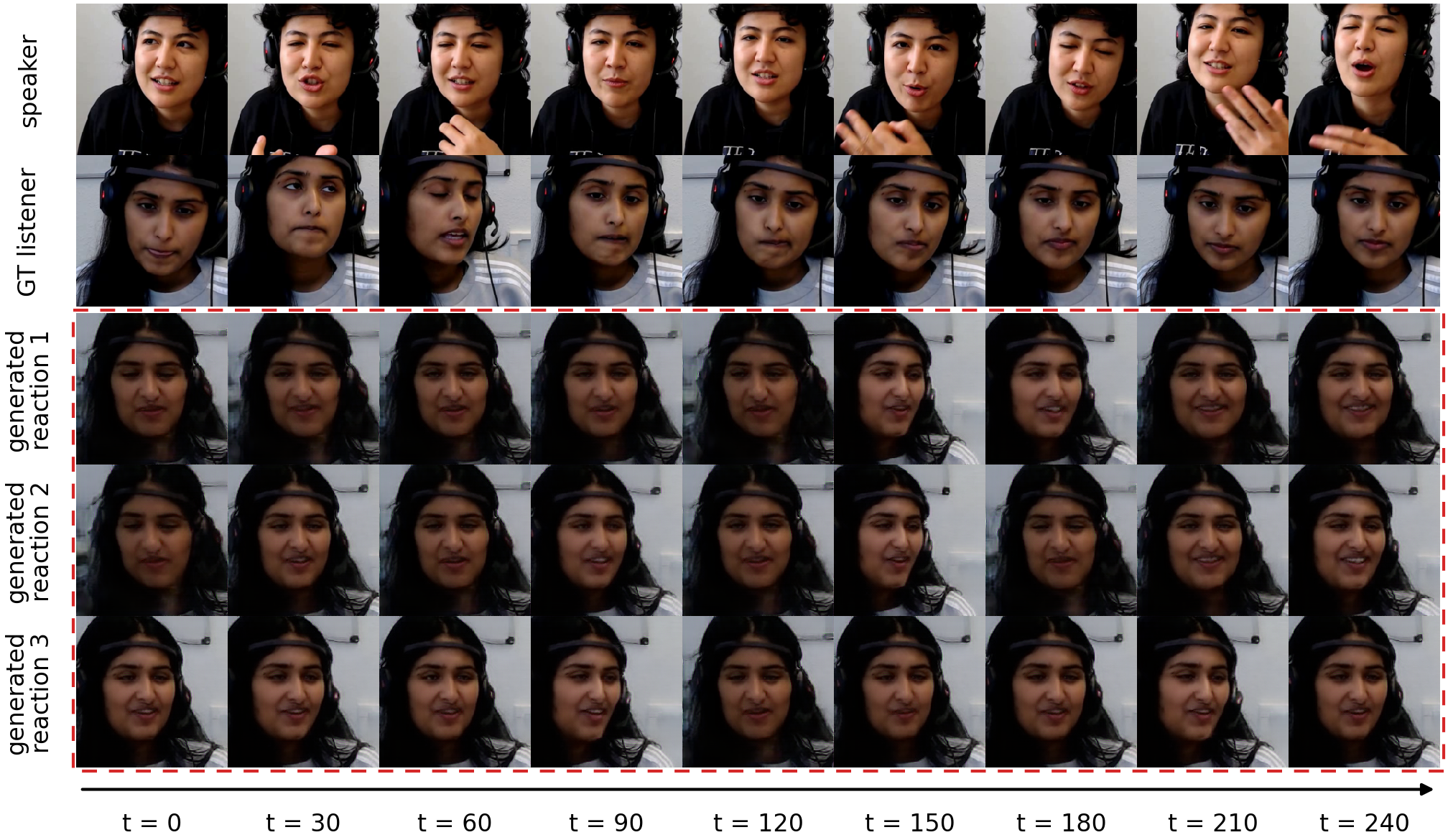}
  \caption{Qualitative reactions generated by ResDiffFRG are presented. The top row shows the input speaker behaviour. The 2nd row shows the corresponding ground-truth listener reaction. The bottom three rows show three independently sampled AFRs conditioned on the same speaker sequence. To visualise the results, each generated $D$-dimensional output is mapped using the latent\_embedder module from PerFRDiff \citep{zhu2024perfrdiff} and then rendered using PIRender. Frames are shown every 30 frames from 0 to 240.}
\label{fig:reaction_figure}
\end{figure}

\end{document}

%% file: math_commands.tex
\usepackage{amsmath,amsfonts,bm}

\def\eqref#1{equation~\ref{#1}}

\def\1{\bm{1}}

\DeclareMathAlphabet{\mathsfit}{\encodingdefault}{\sfdefault}{m}{sl}
\SetMathAlphabet{\mathsfit}{bold}{\encodingdefault}{\sfdefault}{bx}{n}

